\documentclass[useAMS,usenatbib]{coin}
\def\bSig\mathbf{\Sigma}

\newcommand{\ebf}{\it}
\newcommand{\ch}{}

\usepackage[figuresright]{rotating}

 \usepackage{latexsym}
 \usepackage{graphics}
 \usepackage{graphicx}
 \usepackage{latexsym}
 \usepackage{enumerate}
 \title[Crowdsourcing a Word--Emotion Association Lexicon]
 {Crowdsourcing a Word--Emotion Association Lexicon}

 \author{{\sc Saif M. Mohammad and Peter D. Turney}\\
  Institute for Information Technology,
    National Research Council Canada.\\
	 Ottawa, Ontario, Canada, K1A 0R6\\
	    \{saif.mohammad,peter.turney\}@nrc-cnrc.gc.ca}
\begin{document}

% For journal 
\pagerange{\pageref{firstpage}--\pageref{lastpage}} \pubyear{2011}

% For journal 
\volume{59}
% \artmonth{December}
% \doi{10.1111/j.1541-0420.2005.00454.x}

%  This label and the label ``lastpage'' are used by the \pagerange
%  command above to give the page range for the article

\label{firstpage}

%  pub the summary here

\begin{abstract}

Even though considerable attention has been given to the polarity of
words (positive and negative) and the creation of large polarity
lexicons, research in emotion analysis has had to rely on limited and
small emotion lexicons.  In this paper we show how the combined
strength and wisdom of the crowds can be used to generate a large,
high-quality, word--emotion and word--polarity association lexicon
quickly and inexpensively.  We enumerate the challenges in emotion
annotation in a crowdsourcing scenario and propose solutions to
address them.  Most notably, in addition to  questions about emotions
associated with terms, we show how the inclusion of a word choice
question can discourage malicious data entry, help identify instances
where the annotator may not be familiar with the target term (allowing
us to reject such annotations), and help obtain annotations at sense
level (rather than at word level). 
% We perform an extensive analysis of
% the annotations to better understand the distribution of emotions
% associated with terms of different parts of speech.  
% We show that many terms are associated not just with one, but with multiple emotions.
We conducted experiments on how to formulate the emotion-annotation
questions, and show that asking if a term is {\it associated} with an
emotion leads to markedly higher inter-annotator agreement than that
obtained by asking if a term {\it evokes} an emotion.
% The lexicon, with close to 10,000 entries (one entry per word--sense
% pair), will be made publicly available. 

% \thanks{{\tiny (a)
 % 		Guest editors:  Diana Inkpen and Carlo Strapparava.
 % 		(b) Name of conference or workshop of the initial publication:
 % 		NAACL-HLT 2010 Workshop on Computational Approaches to
 % 		Analysis and Generation of Emotion in Text, June 5, 2010, Los
 % 		Angeles, CA.
 % 		(c) The title of the special issue: Computational Approaches
 % 		to Analysis of Emotion in Text.}}
\end{abstract}

%
%  Please place your key words in alphabetical order, separated
%  by semicolons, with the first letter of the first word capitalized,
%  and a period at the end of the list.
%


\maketitle
\vspace*{-12pt}
\section{Introduction}
\label{s:intro}

%  by \citep{ReferenceKey1}. 

%We live in a world where we have to deal with more information every day. 
We call upon computers and algorithms to
assist us in sifting through enormous amounts of data and also to
understand the content---for example, ``What is being said about a
certain target entity?" (Common target entities include a company, product, policy, person, and country.) 
Lately, we are going further, and also asking questions such
as: ``Is something good or bad being said about the target entity?"
and ``Is the speaker happy with, angry at, or fearful of the target?".
This is the area of {\ebf sentiment analysis}, which involves determining the
opinions and private states (beliefs, feelings, and speculations) of
the speaker towards a target entity \cite[]{Wiebe94}.  Sentiment
analysis has a number of applications, for example in managing customer
relations, where an automated system may transfer an angry,
agitated caller to a higher-level manager. An increasing number of companies
want to automatically track the response to their product (especially when there are new
releases and updates) on blogs, forums, social networking sites such
as Twitter and Facebook, and the World Wide Web in general. 
% It allows them to quickly discover
% flaws, new requirements, and positive reviews of their product so that
% they can take appropriate corrective actions, make improvements, and
% identify what is working well, thereby improving consumer
% satisfaction.  
(More applications listed in Section~\ref{sec: apps}.)
Thus, over the last decade,
there has been considerable work in sentiment analysis, and especially in determining whether a word, phrase, or document has a {\ebf positive
polarity}, that is, it is expressing a favorable sentiment towards an entity, or whether it has a {\ebf negative
polarity}, that is, it is expressing an unfavorable sentiment towards an entity
\cite[]{Lehrer74,Turney03,PangL08}. (This sense of {\it polarity} is also referred to as {\ebf semantic orientation} \ch{and} {\it valence}
% , and the two are used interchangeably 
in the literature.) 
However, much research remains to be done on the problem of
automatic analysis of {\ebf emotions} in text.

Emotions are often expressed through
different facial expressions \cite[]{Aristotle13,Russell94}.  
Different emotions are also expressed through different words.  For
example, {\it delightful} and {\it yummy} indicate the emotion of joy,
{\it gloomy} and {\it cry} are indicative of sadness, {\it shout} and
{\it boiling} are indicative of anger, and so on. 
% Over time, a number
% of definitions have been provided for {\it emotion}.
% (\cite{Plutchik80} provides twenty-eight definitions of emotion in his
% book.)  
% {\it Merriam Webster} defines {\it emotion} as follows: 
% \begin{lquote} {\it A
% mental reaction subjectively experienced as strong feeling usually
% directed toward a specific object and typically accompanied by
% physiological and behavioral changes in the body.  }
% \end{lquote}
% \noindent 
In this paper, we are interested in how emotions manifest themselves
in language through words.\footnote{This paper
expands on work first published in \cite{MohammadT10}.}
We describe an  annotation
project aimed at creating a large lexicon of term--emotion
associations. A term is either a word or a phrase.  Each entry in this
lexicon includes a term, an
emotion, and a measure of how strongly the term is associated with the emotion.  Instead of providing definitions for the
different emotions, we give the
annotators examples of words associated with different emotions and rely on their intuition of
what different emotions mean and how language is used to express
emotion.

Terms may evoke different emotions in different contexts, and the
emotion evoked by a phrase or a sentence is not simply the sum of
emotions conveyed by the words in it. However, the emotion lexicon can be
a useful component for a sophisticated emotion detection algorithm
required for many of the applications described in the next section.
The term--emotion association lexicon will also be useful for
evaluating automatic methods that identify the emotions associated with a
word. Such algorithms may then be used to automatically generate
emotion lexicons in languages where no such lexicons exist.  As of
now, high-quality, high-coverage, emotion lexicons do not exist for
any language, although there are a few limited-coverage lexicons for a
handful of languages, for example, the WordNet Affect Lexicon (WAL)
\cite[]{StrapparavaV04}, the General
Inquirer (GI) \cite[]{Stone66}, and the \ch{Affective Norms} for English Words (ANEW)
\cite[]{BradleyL99}. 
%, which categorizes words into a number
% of categories, including positive and negative semantic orientation.

The lack of emotion resources can be attributed to high cost and
considerable manual effort required of the human annotators in a
traditional setting where hand-picked experts are hired to do all the annotation.  
However, lately a new model
has evolved to do large amounts of work quickly and inexpensively.
{\ebf Crowdsourcing} is the act of breaking down work into many small
independent units and distributing them to a large number of people,
usually over the web.  
\cite{HoweR06}, who coined the term,
define it as follows:\footnote{http://crowdsourcing.typepad.com/cs/2006/06}
\begin{lquote}
{\it The act of a company or institution taking a function once performed
by employees and outsourcing it to an undefined (and generally large)
network of people in the form of an open call. This can take the form
of peer-production (when the job is performed collaboratively), but is
also often undertaken by sole individuals. The crucial prerequisite is
the use of the open call format and the large network of potential
laborers.}
\end{lquote}
\noindent Some well-known crowdsourcing projects include Wikipedia,
Threadless, iStockphoto, InnoCentive, Netflix Prize, and Amazon's
Mechanical Turk.\footnote{ Wikipedia: http://en.wikipedia.org,
Threadless: http://www.threadless.com,\\ iStockphoto:
http://www.istockphoto.com,  InnoCentive:
http://www.innocentive.com,\\ Netflix prize:
http://www.netflixprize.com, Mechanical Turk:
https://www.mturk.com/mturk/welcome}

% Crowdsourcing, as defined by Wikipedia (itself arguably the most well-known crowdsourcing project), is {\it ``the act of 
% outsourcing tasks, traditionally performed by an employee or contractor, 
% to a large group of people or community (a crowd), through an open call."}

Mechanical Turk is an online crowdsourcing platform that is especially
suited for tasks that can be done over the Internet through a computer
or a mobile device.  It is already being used to obtain human annotation
on various linguistic tasks \cite[]{SnowOJN08,CallisonBurch09}.
However, one must define the task carefully to obtain annotations of
high quality. Several checks must be placed to ensure that random and
erroneous annotations are discouraged, rejected, and re-annotated.

In this paper, we show how we compiled a large English term--emotion association lexicon
by manual annotation through Amazon's Mechanical Turk service.  This
dataset, which we call {\ebf EmoLex}, is an order of magnitude
larger than the WordNet Affect
Lexicon.  We focus on the emotions of joy, sadness, anger, fear,
trust, disgust, surprise, and anticipation---argued by many to be the
basic and prototypical emotions \cite[]{Plutchik80}.  The terms in
EmoLex are carefully chosen to include some of the most frequent
English nouns, verbs, adjectives, and adverbs. In addition to unigrams, EmoLex has many
commonly used bigrams as well.  We also include words from the General
Inquirer and the WordNet Affect Lexicon to allow comparison of
annotations between the various resources.  We perform extensive
analysis of the annotations to answer several questions, including the following:
\begin{enumerate}[1.]
\item How hard is it for humans to annotate words with their associated emotions?
\item How can emotion-annotation questions be phrased to make them
accessible and clear to the average English speaker?
\item Do small differences in how the questions are asked 
% (for example, ``is the emotion evoked" or ``is the emotion associated") 
result in significant annotation differences?
% \item What percentage of commonly used terms, in each part of speech,
% evoke an emotion?  
\item Are emotions more commonly evoked by nouns, verbs,
adjectives, or adverbs? How common are emotion terms among the various parts of speech?
\item How much do people agree on the association of a given emotion with a given word?
\item Is there a correlation between the
polarity of a word and the emotion associated with it? 
\item Which emotions tend to go together; that is, which emotions are associated
with the same terms?  
\end{enumerate}
% What colours are associated with different emotions?  What percentage of words have strong colour
% associations?  What colours are most commonly associated with words?
% And so on.  
\noindent Our lexicon now has close to 10,000 terms and ongoing work will 
make it even larger (we are aiming for about 40,000 terms).

\section{Applications}
\label{sec: apps}

% The term--emotion lexicon (and emotion analysis in general)
\ch{The automatic recognition of emotions}
is useful for a number of tasks, including the following:
\begin{enumerate}[1.]
\item Managing customer relations
by taking appropriate actions depending on the customer's emotional state (for example, dissatisfaction, satisfaction, sadness,
trust, anticipation, or anger) \cite[]{BougiePZ03}.  

\item Tracking sentiment towards politicians, movies, products,
countries, and other target entities \cite[]{PangL08,MohammadY11}.
\item Developing sophisticated search algorithms that distinguish
between different emotions associated with a product
\cite[]{KnautzSS10}. For example, customers may search for banks,
mutual funds, or stocks that people trust. Aid
organizations may search for events and stories that are generating 
empathy, and highlight them in their fund-raising campaigns.
Further, systems that are not emotion-discerning may fall prey to
abuse. For example, it was recently discovered that an online vendor
deliberately mistreated his customers because the negative online reviews
translated to higher rankings on
Google \ch{searches}.\footnote{http://www.pcworld.com/article/212223/google\_algorithm\_will\_punish\_bad\_businesses.html}
\item Creating dialogue systems that respond appropriately to
different emotional states of the user; for example, in 
emotion-aware games \cite[]{Velasquez97,RavajaSTLSK06}.
\item Developing intelligent tutoring systems that manage the emotional state
of the learner for more effective learning. \ch{There} is some support for the hypothesis that
students learn better and faster when they are in a positive emotional state \cite[]{LitmanF04}.
\item Determining risk of repeat attempts by analyzing suicide notes \cite[]{Osgood59,Matykiewicz09,Pestian08}.\footnote{The 2011 Informatics for Integrating Biology and the Bedside (i2b2) challenge by
the National Center for Biomedical Computing is on detecting emotions in suicide notes.}
\item Understanding how genders communicate through work-place and personal email \cite[]{MohammadY11}.
\item Assisting in writing e-mails, documents, and other text to convey the
desired emotion (and avoiding misinterpretation) \cite[]{LiuLS03}.
\item Depicting the flow of emotions in novels and other books \cite[]{Boucouvalas02,Mohammad11b}.
\item Identifying what emotion a newspaper headline is trying to
evoke \cite[]{Bellegarda10}.
\item Re-ranking and categorizing information/answers in online question--answer forums
\cite[]{AdamicZBEM08}. For example, highly emotional responses may be ranked lower.
\item Detecting how people use emotion-bearing-words and metaphors to
persuade and coerce others (for example, in propaganda) \cite[]{Zoltan03}.
\item Developing more natural text-to-speech systems \cite[]{FranciscoG06,Bellegarda10}.
\item Developing assistive robots that are sensitive to human emotions
\cite[]{BreazealB04,HollingerGMMPM06}. For example, the robotics group
in Carnegie Melon University is interested in building an
emotion-aware physiotherapy coach robot.
\end{enumerate}

Since we do not have space to fully explain all of these
applications, we select one (the first application from the list: managing customer relations)
to develop in more detail as an illustration of the value of emotion-aware systems.
% \item Determining emotional intelligence by analyzing documents written by a person.
\cite{Davenport01} define {\it customer relationship management (CRM)} systems as:
\begin{lquote}
{\it All the tools, technologies
and procedures to manage, improve or facilitate sales, support and
related interactions with customers, prospects, and business partners
throughout the enterprise.}
\end{lquote}
\noindent Central to this process is keeping the
customer satisfied. A number of studies have looked at dissatisfaction
and anger and shown how they can lead to complaints to company
representatives, litigations against the company in courts, {\ebf negative
word of mouth}, and other outcomes that are detrimental to company goals
\cite[]{MauteF93,Richins87,Singh88}.  
\cite{Richins84} defines {\it negative word of mouth} as:
\begin{lquote}
{\it Interpersonal communication among consumers concerning a marketing organization or product which denigrates the object of the communication.}
\end{lquote}
\noindent Anger, as indicated earlier, is
clearly an emotion, and so is dissatisfaction
\cite[]{OrtonyCC88,Scherer84,ShaverSKO87,Weiner85}.  Even though the
two are somewhat correlated \cite[]{FolkesKG87}, \cite{BougiePZ03}
show through experiments and case studies that dissatisfaction and
anger are distinct emotions, leading to distinct actions by the
consumer.  Like \cite{Weiner85}, they argue that dissatisfaction is
an ``outcome-dependent emotion", that is, it is a reaction to an
undesirable outcome of a transaction, and that it instigates the
customer to determine the reason for the undesirable outcome. If
customers establish that it was their own fault, then this may evoke
an emotion of guilt or shame.  If the situation was beyond anybody's
control, then it may evoke sadness. However, if they feel that it was
the fault of the service provider, then there is a tendency to become
angry.  Thus, dissatisfaction is usually a precursor to anger (also
supported by \cite{Scherer82,Weiner85}), but may often instead lead to
other emotions such as sadness, guilt, and shame, too.
\cite{BougiePZ03} also show that dissatisfaction does not have a
correlation with complaints and negative word of mouth, when \ch{the data} is
controlled for anger.  
On the other hand, anger has a strong
correlation with complaining and negative word of mouth, even when
satisfaction is controlled for \cite[]{DiazR02,DubeM96}.

Consider a scenario in which a company has automated systems on the
phone and on the web to manage high-volume calls. Basic queries and
simple complaints are handled automatically, but non-trivial ones are
forwarded to a team of qualified call handlers. It is
usual for a large number of customer interactions to have negative
polarity terms because, after all, people often contact a company
because they are dissatisfied with a certain outcome. However, if the
system is able to detect that a certain caller is angry (and thus, if
not placated, is likely to engage in negative word of mouth about the
company or the product), then it can immediately transfer the call to
a qualified higher-level human call handler.  

Apart from keeping the customers satisfied, companies are also interested
in developing a large base of loyal customers. Customers loyal to a company buy more
products, spend more money, and also spread positive word of mouth \cite[]{HarrisG04}.
\cite{Oliver97}, \cite{DabholkarST00}, \cite{HarrisG04}, and others give evidence that central to attaining
loyal customers is the amount of trust they have in the company.
Trust is especially important in on-line services
where it has been shown that consumers buy more and return more often to
shop when they trust a company
\cite[]{ShankarUS02,ReichheldS00,Stewart03}.
% argue that it is important for a company to maintain trust in all
%  modes of interaction with the consumer, whether it is through the store front,
%  online, or automated machines such as ATMs.

Thus it is in the
interest of the company to heed the consumers, not just when
they call, but also during online transactions and when they write about the company in their blogs,
tweets, consumer forums, and review websites so that they can
immediately know whether the customers are happy with,
dissatisfied with, losing  trust in, or angry with their product or a particular feature of
the product. This way they can take corrective action when necessary,
and accentuate the most positively evocative features.
Further, an emotion-aware system can discover instances of high trust and 
use them as sales opportunities (for example, offering a related product or service for purchase).

\section{Emotions}
Emotions are pervasive among humans, and \ch{many} are innate. Some argue that
even across cultures that have no contact with each other, facial
expressions for basic human emotions are identical
\cite[]{EkmanF03,Ekman05}.  
% They also suggest that there may be a link between perceiving the basic
% emotions and the associated facial expressions \cite[]{DePaulo92}.
However, other studies argue that 
there may be some universalities, but language and culture play an
important role in shaping our emotions and also in how they manifest
themselves in facial expression \cite[]{ElfenbeinA02,Russell94}.
There is some contention on whether animals have emotions, but there
are studies, especially for higher mammals, canines, felines, and even
some fish, arguing in favor of the proposition
\cite[]{Masson96,GuoHMM07}.  Some of the earliest work is by Charles
Darwin in his book {\it The Expressions of the Emotions in Man and
Animals} \cite[]{Darwin72}. Studies by evolutionary biologists and
psychologists show that emotions have evolved to improve the
reproductive fitness for a species, as they are triggers for
behavior with high survival value.  For example, fear inspires
fight-or-flight response.  The more complex brains of primates
and humans are capable of experiencing not just the basic
emotions such as fear and joy, but also more complex and nuanced
emotions such as optimism and shame.
% http://en.wikipedia.org/wiki/Emotion_in_animals
% In this study, we are interested in human emotions associated with words. 
% There are at least a few hundred human emotions and e 
Similar to emotions, other phenomena such as {\ebf mood}
also pertain to the evaluation of one's well-being and are together
referred to as {\ebf affect}
\cite[]{Scherer84,Gross98,Steunebrink10}.
% The use of the term {\bf affect} varies across different
% psychologists, but most consider it to be a super-ordinate category,
% that includes not just emotions but also {\bf mood} and certain other
% categories, all of which deal with the evaluation of one’s well-being
% \cite[]{Scherer84,Gross98,Steunebrink10}.  
Unlike emotion, mood is not
towards a specific thing, but more diffuse, and it lasts for longer
durations
\cite[]{NowlisN56,Gross98,Steunebrink10}.

Psychologists have proposed a number of theories that classify 
% the hundreds of emotions that humans can experience 
\ch{human emotions} into taxonomies.
As mentioned earlier, some emotions are considered basic, whereas
others are considered complex.  
Some psychologists
have classified emotions into those that we can sense and perceive
({\ebf instinctual}), and those that that we arrive at after some
thinking and reasoning ({\ebf cognitive}) \cite[]{Zajonc84}.  However,
others do not agree with such a distinction and argue that emotions do
not precede cognition \cite[]{Lazarus84,Lazarus00}.
\cite{Plutchik85} argues that this debate may not be resolvable
because it does not lend itself to empirical proof and that the
problem is a matter of definition.
There is a high correlation between the basic and
instinctual emotions, as well as between complex and cognitive
emotions.  Many of the basic emotions are also instinctual.

% Instinctual emotions have been associated with the amygdala 
% and are found even in many of the species with a very primitive brain.
% Cognitive emotions are associated with the pre-frontal cortex and found in
% species with more developed brains. 
% (Figure~\ref{fig:amygdala} shows the amygdala and pre-frontal cortex in a human brain.) 

%  \begin{figure}[t]
%  \begin{center}
%  \includegraphics[width=0.55\columnwidth]{amygdala.png}
%  \end{center}
%  \caption{Illustration of the human brain showing the amygdala (associated with instinctual emotions)
%  and the pre-frontal cortex (associated with cognitive emotions).}
%  \label{fig:amygdala}
%  \end{figure}

 % http://www.google.co.in/imgres?imgurl=http://thesituationist.files.wordpress.com/2007/06/amygdala.jpg&imgrefurl=http://thesituationist.wordpress.com/2007/06/27/a-penny-for-your-thoughts-may-be-the-deal-of-the-century/&h=1133&w=1200&sz=121&tbnid=LsBW3RXSH0tzhM:&tbnh=142&tbnw=150&prev=/images%3Fq%3Damygdala&usg=__cTZZg4GqA1csT6eXOhzaDF9ZZCk=&sa=X&ei=x7JSTNPOFouwrAfsofD2Bg&ved=0CC8Q9QEwAw

% Since annotating for hundreds of emotions is expensive and also hard for annotators,
% we annotate the words with only the basic emotions.

\begin{figure}[t]
\begin{center}
\includegraphics[width=0.6\columnwidth]{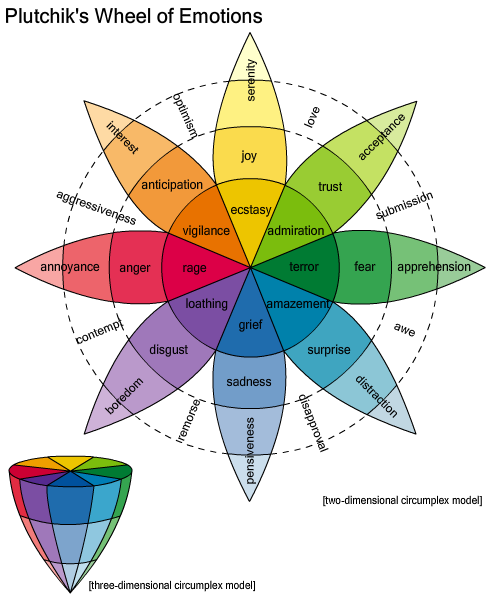}
\end{center}
\caption{Plutchik's wheel of emotions. Similar emotions are placed next to each other. Contrasting
emotions are placed diametrically opposite to each other. Radius indicates intensity.
White spaces in between the basic emotions represent primary dyads---complex emotions
that are combinations of adjacent basic emotions. (The image file is taken from Wikimedia Commons.)}
\label{fig:plutchik wheel}
\end{figure}

A number of theories have been proposed on which
emotions are basic \cite[]{Ekman92,Plutchik62,Parrot01,James84}.  See
\cite{OrtonyT90} for a detailed review of many of these models.
\cite{Ekman92} argues that there are six basic emotions: joy, sadness,
anger, fear, disgust, and surprise.
\cite{Plutchik62,Plutchik80,Plutchik94} proposes a theory with eight
basic emotions. These include Ekman's six as well as trust and
anticipation.  Plutchik organizes the emotions in a wheel (Figure~\ref{fig:plutchik wheel}).  The radius indicates intensity---the
closer to the center, the higher the intensity.  
% The Plutchik wheel is good at visualizing what he argues are opposing emotions by placing
% them diametrically opposite to each other (joy--sadness, anger--fear,
% trust--disgust, and anticipation--surprise).  
Plutchik argues that the eight basic emotions form four opposing pairs,
joy--sadness, anger--fear, trust--disgust, and anticipation--surprise.
This emotion opposition is displayed in Figure~\ref{fig:plutchik wheel} by the
spatial opposition of these pairs.
The figure also shows certain
emotions, called {\ebf primary dyads}, in the white spaces between the basic emotions, which he
argues can be thought of as combinations of the adjoining emotions.
However it should be noted that emotions in general do not have clear
boundaries and do not always occur in isolation.  
% (Further, the use of
% basic colours to represent different basic emotions is probably just
% incidental, he does not propose a correlation between emotions and
% their colours in the wheel.)

% \begin{minipage}[t]{0.5\textwidth}
% \end{minipage}

Since annotating words with hundreds of emotions is expensive for us
and \ch{difficult} for annotators, we decided to annotate words with \ch{Plutchik's} eight 
basic emotions.  
% We chose the eight basic emotions proposed by
% Plutchik not because we are certain that these are the true basic
% emotions, but because: 
We do not claim that Plutchik's eight emotions are more fundamental than other categorizations; 
however, we adopted them for annotation purposes because:
(a) like some of the other choices of basic
emotions, this choice too is well-founded in psychological,
physiological, and empirical \ch{research}, (b) unlike some other choices, for
example that of Ekman, it is not composed of mostly negative emotions,
(c) it is a superset of the emotions proposed by some others (for
example, it is a superset of Ekman's six basic emotions), and 
(d) in our future work, we will conduct new annotation experiments
to empirically verify whether certain pairs of these emotions 
are indeed in opposition or not, and whether the primary dyads can indeed
be thought of as combinations of the adjacent basic emotions.

% \cite[]{LinZQZ03,MohammadDH08,LobanovaKS10}.

% However, it should be noted that there are many other models of basic emotions
% besides it and the one proposed by
% Ekman, such as that of \cite{Parrot01} and \cite{James84}.
% See \cite{OrtonyT90} for a detailed review of many of these models.

% See http://changingminds.org/explanations/emotions/basic%20emotions.htm
% definition; subjective; strong; directed towards
% our goal

% % \begin{minipage}[t]{0.5\textwidth}
% \begin{figure}[t]
% \begin{center}
% \includegraphics[width=0.6\columnwidth]{Plutchik_dyader1.jpg}
% \end{center}
% \caption{Robert Plutchik's set of primary dyads, secondary dyads, tertiary dyads, and opposites.
% Primary Dyads are formed by combinations of adjacent basic emotions.
% Secondary Dyads are formed by combinations of basic emotions that are one hop away on the emotion wheel.
% Tertiary Dyads are formed by combinations of basic emotions that are two hops away on the emotion wheel.
% }
% \label{fig:dyader}
% \end{figure}
% % \end{minipage}

\section{Related work}

 % Emotion detection and analysis is gaining more attention in recent years
 % as there is more and more emotion-rich content on the World Wide Web, for example
 % through blogs and product/service reviews. 

Over the past decade, there has been a large amount of work on sentiment analysis that
focuses on positive and negative polarity. 
\cite{PangL08} provide an excellent summary. Here we focus on
the relatively small amount of work on generating emotion lexicons and
on computational analysis of the emotional content of text.

The WordNet Affect Lexicon (WAL) \cite[]{StrapparavaV04} has a few hundred words
annotated with the emotions they
evoke.\footnote{http://wndomains.fbk.eu/wnaffect.html} It was created
by manually identifying the emotions of a few seed words and then
marking all their WordNet synonyms as having the same emotion. 
The words in WAL are annotated for a number of emotion and affect categories, but its creators
also provided a subset corresponding to the six Ekman emotions.
In our Mechanical Turk experiments, we re-annotate hundreds of words from the Ekman subset of WAL to determine how much 
the emotion annotations obtained from untrained \ch{volunteers} matches that obtained from the original hand-picked judges  (Section 10).
General Inquirer (GI) \cite[]{Stone66} has 11,788 words labeled with 182
categories of word tags, including positive and negative semantic
orientation.\footnote{http://www.wjh.harvard.edu/$\sim$inquirer} It
also has certain other affect categories, such as pleasure, arousal,
feeling, and pain, but these have not been exploited to a significant
degree by the natural language processing community.
In our Mechanical Turk experiments, we re-annotate thousands of words from GI to determine how much
the polarity annotations obtained from untrained \ch{volunteers} matches that obtained from the original hand-picked judges  (Section 11).
\ch{Affective Norms} for English Words (ANEW) has pleasure (happy--unhappy), arousal (excited--calm), and dominance (controlled--in control) ratings
for 1034 words.\footnote{http://csea.phhp.ufl.edu/media/anewmessage.html}

Automatic systems for analyzing emotional content of text follow
many different approaches: a number of these systems look for specific emotion
denoting words \cite[]{Elliott92}, some determine the tendency of terms to
co-occur with seed words whose emotions are known \cite[]{Read04},
some use hand-coded rules \cite[]{NeviarouskayaPI09,Neviarouskaya10}, and some use
machine learning and a number of emotion features, including emotion
denoting words \cite[]{AlmRS05,AmanS07}. Recent work by
\cite{Bellegarda10} uses sophisticated dimension reduction techniques
(variations of latent semantic analysis), to automatically identify
emotion terms, and obtains marked improvements in classifying
newspaper headlines into different emotion categories.
\cite{GoyalRDG10} move away from classifying sentences from the
writer's perspective, towards attributing mental states to entities
mentioned in the text. Their work deals with polarity,
but work on attributing emotions to entities
mentioned in text is, similarly, a promising area of future work.

Much recent work focuses on six emotions studied by
\cite{Ekman92} and \cite{SauteraEES10}. These emotions---joy, sadness,
anger, fear, disgust, and surprise---are a subset of the eight
proposed in \cite{Plutchik80}.  There is less work on complex
emotions, for example, work by \cite{PearlS10} that focuses on
politeness, rudeness, embarrassment, formality, persuasion, deception,
confidence, and disbelief.  They developed a game-based annotation
project for these emotions.  \cite{FranciscoG06} marked sentences in
fairy tales with tags for pleasantness, activation, and dominance,
using lexicons of words associated with the three categories.

Emotion analysis can be applied to all kinds of text, but 
certain domains and modes of communication tend have more overt expressions of emotions than
others.  \cite{Neviarouskaya10}, \cite{GenereuxE06}, and \cite{MihalceaL06} analyzed web-logs.
\cite{AlmRS05} and \cite{FranciscoG06} worked on fairy tales.
\cite{Boucouvalas02} and \cite{JohnBX06} explored emotions in novels.
\cite{ZheB02}, \cite{HolzmanP03}, and \cite{MaPI05} annotated chat
messages for emotions.  \cite{LiuLS03} worked on email data.

There has also been some interesting work in visualizing emotions, for
example that of \cite{SubasicH01}, \cite{KalraK05}, and
\cite{RashidAF06}. \cite{Mohammad11a} describes work on identifying colours 
associated with emotion words.

\section{Target terms}
\label{sec:terms}

In order to generate a word--emotion association lexicon, we  first
identify a list of words and phrases for which we want human
annotations. We chose the {\it Macquarie Thesaurus} as our source 
for unigrams and bigrams \cite[]{Bernard86}.\footnote{http://www.macquarieonline.com.au/thesaurus.html}
The categories in the thesaurus act as
coarse senses of the words. (A word listed in two categories is taken
to have two senses.) Any other published dictionary would have worked
well too.  Apart from over 57,000 commonly used English word
types, the {\it Macquarie Thesaurus} also has entries for more than
40,000 commonly used phrases.  From this list we chose those terms
that occurred frequently in the Google n-gram corpus
\cite[]{BrantsF06}. Specifically we chose the 200 most frequent
unigrams and 200 most frequent bigrams from four parts of speech:
nouns, verbs, adverbs, and adjectives.  When selecting these sets, we
ignored terms that occurred in more than one {\it Macquarie Thesaurus}
category.  (There were only 187 adverb bigrams that matched these
criteria. All other sets had 200 terms each.) 
We chose all words from
the Ekman subset of the WordNet Affect Lexicon that
had at most two senses (terms listed in at most two thesaurus
categories)---640 word--sense pairs in all.  We included all terms in
the General Inquirer that were not too ambiguous (had at most three 
senses)---8132 word--sense pairs in all. (We started the
annotation on monosemous terms, and gradually included more 
ambiguous terms as we became confident that the quality
of annotations was acceptable.) Some of these terms occur in
more than one set.  The union of the three sets (Google n-gram terms,
WAL terms, and GI terms) has 10,170 term--sense pairs.
Table~\ref{tab:target terms} lists the various
 sets of target terms as well as the number of terms in each set for which
 annotations were requested.
 EmoLex-Uni stands for all the unigrams taken from
 the thesaurus. EmoLex-Bi refers to all the bigrams taken from the thesaurus.
EmoLex-GI are all the words taken from the General Inquirer.
EmoLex-WAL are all the words taken from the WordNet Affect Lexicon.

\begin{table}[t]
\caption{Break down of the target terms for which emotion annotations were requested.} 
\centering
% \resizebox{0.5\textwidth}{!}{
\begin{tabular}{lrr}
\hline
						
	{\bf EmoLex} 		&{\bf \# of terms}    &{\bf \% of the Union}\\
\hline
	 % \multicolumn{3}{l}{\bf $\textrm{EmoLex}_\textrm{Uni}$:} \\
	\multicolumn{3}{l}{\bf EmoLex-Uni:} \\
	\multicolumn{3}{l}{Unigrams from Macquarie Thesaurus}  \\
	$\;\;\;\;$ adjectives      &200  &2.0\%\\
	$\;\;\;\;$ adverbs          &200 &2.0\%\\
	$\;\;\;\;$ nouns            &200 &2.0\%\\
	$\;\;\;\;$ verbs            &200 &2.0\%\\

	 % \multicolumn{3}{l}{\bf $\textrm{EmoLex}_\textrm{Bi}$: \rule{0pt}{11pt}} \\
	\multicolumn{3}{l}{\bf EmoLex-Bi: \rule{0pt}{11pt}} \\
	\multicolumn{3}{l}{Bigrams from Macquarie Thesaurus}  \\
	$\;\;\;\;$ adjectives       &200 &2.0\%\\
	$\;\;\;\;$ adverbs          &187 &1.8\% \\
	$\;\;\;\;$ nouns            &200 &2.0\% \\
	$\;\;\;\;$ verbs            &200 &2.0\%\\

	 % \multicolumn{3}{l}{\bf $\textrm{EmoLex}_\textrm{GI}$: \rule{0pt}{11pt}}  \\
	\multicolumn{3}{l}{\bf EmoLex-GI: \rule{0pt}{11pt}} \\
\multicolumn{3}{l}{Terms from General Inquirer}  \\
	$\;\;\;\;$ negative terms          &2119  &20.8\%\\
	$\;\;\;\;$ neutral terms           &4226  &41.6\%\\
	$\;\;\;\;$ positive terms          &1787  &17.6\% \\

	 % \multicolumn{3}{l}{\bf $\textrm{EmoLex}_\textrm{WAL}$: \rule{0pt}{11pt}} \\
	\multicolumn{3}{l}{\bf EmoLex-WAL: \rule{0pt}{11pt}} \\
	\multicolumn{3}{l}{Terms from WordNet Affect Lexicon} \\
	$\;\;\;\;$ anger terms           &165  &1.6\%\\
	$\;\;\;\;$ disgust terms        &37   &0.4\%\\
	$\;\;\;\;$ fear terms           &100  &1.0\%\\
	$\;\;\;\;$ joy terms             &165 &1.6\% \\
	$\;\;\;\;$ sadness terms        &120  &1.2\%\\
	$\;\;\;\;$ surprise terms       &53   &0.5\%\\
	{\bf Union}	\rule{0pt}{10pt}								&{\bf 10170} &{\bf 100\%}\\
\hline
\\
\end{tabular}
% }
\label{tab:target terms}
\end{table}

% Words in General Inquirer have been annotated into a number of classes including whether
% they have positive or negative semantic orientation.
% The first and second column of Table~\ref{tab:emotion terms} list the various
% sets of target terms as well as the number of terms in each set for which
% annotations were requested.
% {\bf $\textrm{EmoLex}_\textrm{Uni}$} stands for all the unigrams taken from
% the thesaurus. {\bf $\textrm{EmoLex}_\textrm{Bi}$} refers to all the bigrams.
% {\bf $\textrm{EmoLex}_\textrm{GI}$} are all the words taken from the General Inquirer.
% {\bf $\textrm{EmoLex}_\textrm{WAL}$} are all the words taken from the WordNet Affect Lexicon.

\section{Mechanical Turk}

We used Amazon's Mechanical Turk service as a platform to obtain
large-scale emotion annotations.  An entity submitting a task to
Mechanical Turk is called the {\ebf requester}.  The requester breaks
the task into small independently solvable units called {\ebf HITs
(Human Intelligence Tasks)} and uploads them on the Mechanical Turk
website.  The requester specifies (1) some key words relevant to the
task to help interested people find the HITs on Amazon's website, (2)
the compensation that will be paid for solving each HIT, and (3) the
number of different annotators that are to solve each HIT.  The people
who provide responses to these HITs are called {\ebf Turkers}.  Turkers
usually search for tasks by entering key words representative of the
tasks they are interested in and often also by specifying the minimum
compensation per HIT they are willing to work for.  The annotation
provided by a Turker for a HIT is called an {\ebf assignment}.

We created Mechanical Turk HITs for each of the terms specified in
Section~\ref{sec:terms}.  Each HIT has a set of questions, all of
which are to be answered by the same person.  (A complete example HIT
with directions and all questions is shown in Section~\ref{sec:our
approach} ahead.) 
% We requested five different assignments for each HIT (each HIT is to be annotated by five different Turkers).  
We requested annotations from five different Turkers for each HIT. 
(A Turker cannot attempt \ch{multiple} assignments for the same term.)
Different
HITS may be attempted by different Turkers, and a Turker may attempt
as many HITs as they wish.

\section{Issues with crowdsourcing and emotion annotation}
% In the subsections below, we describe the main issues with
% a crowdsourcing platform such as Mechanical Turk, and some of the finer points
% in obtaining emotion annotations.

\subsection{Key issues in crowdsourcing}

Even though there are a number of benefits to using Mechanical Turk, such as
low cost, less organizational overhead, and quick turn around time,
there are also some inherent challenges. First and foremost is quality
control. The task and compensation may attract cheaters (who may input
random information) and even malicious annotators (who may
deliberately enter incorrect information). We have no control over the
educational background of a Turker, and we cannot expect the average
Turker to read and follow complex and detailed directions.
% that one can some times get away with in a more traditional annotation setting.
However, this may not necessarily be a disadvantage of crowdsourcing.
We believe that clear, brief, and simple
instructions produce accurate annotations and higher inter-annotator agreements.  
Another challenge  % in crowdsourcing 
is finding enough
Turkers interested in doing the task. If the task does not require any
special skills, then more Turkers will do the task.  The number of
Turkers and the number of annotations they provide is also dependent
on how interesting they find the task and how attractive they find the compensation.
% A high compensation is not always the best approach though, since
% it attracts cheaters. With the right amount of compensation one can
% attract Turkers who are doing the task not just for the money but also
% because they find the task interesting.

\subsection{Finer points of emotion annotation}

Native and fluent speakers of a language are good at identifying
emotions associated with words.
% (as we show in Section \ref{sec: emo analysis}). 
Therefore we do not require the
annotators to have any special skills other than that they be native
or fluent speakers of English. However, emotion annotation, especially
in a crowdsource setting, has some important challenges.

Words used in different senses can evoke different emotions.  For
example, the word {\it shout} evokes a different emotion when used in
the context of admonishment than when used in ``{\it Give me a shout
if you need any help.}" Getting human annotations for word senses is
made complicated by decisions about which sense-inventory to use and
what level of granularity the senses must have.  On the one hand, we
do not want to choose a fine-grained sense-inventory because then the
number of word--sense combinations will become too large and difficult
to easily distinguish, and on the other hand we do not want to work
only at the word level because, when used in different senses, a word
may evoke different emotions.

Yet another challenge is how best to convey a word sense to the
annotator.  Including long definitions will mean that the annotators have to spend more
time reading the question, and because their compensation is roughly proportional
to the amount of time they spend on the task,
the number of annotations we can obtain for a given budget is impacted.
Further, we want the users to annotate a word only if they are already
familiar with it and know its meanings.  Definitions are good at
conveying the core meaning of a word but they are not so effective in conveying
the subtle emotional connotations.  Therefore we wanted to discourage
Turkers from annotating for words they are not familiar with.  
Lastly, we must ensure that malicious and erroneous annotations are
discarded.

\section{Our approach}
\label{sec:our approach}
In order to overcome the challenges described above, before asking the
annotators questions about which emotions are associated with a target
term, we first present them with a word choice problem.  They are
provided with four different words and asked which word is closest in
meaning to the target. 
Three of the four options are irrelevant distractors. The remaining 
option is a synonym for one of the senses of the target word.
This single question serves many purposes.
Through this question we convey the word sense for which annotations
are to be provided, without actually providing annotators with long
definitions.  
That is, the correct choice guides the Turkers to the intended sense
of the target.
Further, if an annotator is not familiar with the target
word and still attempts to answer questions pertaining to the target,
or is randomly clicking options in our questionnaire, then there is a
75\% chance that they will get the answer to this question wrong, and
we can discard all responses pertaining to this target term by the
annotator (that is, we also discard answers to the emotion questions
provided by the annotator for this target term).

We generated these word choice problems automatically using the {\it
Macquarie Thesaurus} \cite[]{Bernard86}.  As mentioned earlier in Section~\ref{sec:terms},
published thesauri, such as {\it Roget's} and {\it Macquarie}, divide
the vocabulary into about a thousand categories, which may be
interpreted as coarse senses.  Each category has a head word that
best captures the meaning of the category.  The word choice question
for a target term is automatically generated by selecting the
following four alternatives (choices): the head word of the thesaurus
category pertaining to the target term (the correct answer); and three
other head words of randomly selected categories (the distractors).
The four alternatives are presented to the annotator in random order.
We generated a separate HIT (and a separate word choice question) for
every sense of the target.
% For these questions, we listed head words from both the senses (categories)
% as two of the alternatives (probability of a random choice being correct is 50\%). 
% Depending on the alternative chosen, we can thus
% determine the sense for which the subsequent emotion responses are provided by the annotator.
We created Mechanical Turk HITs for each of the terms (n-gram--sense
pairs) specified in Table~\ref{tab:target terms}.  
% The first and
% second column of Table~\ref{tab:emotion terms} list the various sets
% of target terms as well as the number of terms in each set for which
% annotations were requested.
%  {\bf $\textrm{EmoLex}_\textrm{Uni}$} stands for all the unigram--sense pairs taken from the thesaurus. 
%  {\bf $\textrm{EmoLex}_\textrm{Bi}$} refers to all the bigram--sense pairs.
%  {\bf $\textrm{EmoLex}_\textrm{GI}$} are all the word--sense pairs taken from the General Inquirer.
%  {\bf $\textrm{EmoLex}_\textrm{WAL}$} are all the word--sense pairs taken from the WordNet Affect Lexicon.
Each HIT has a set of questions, all of which are to be answered by
the same person.  
As mentioned before, we requested five independent
assignments (annotations) for each HIT.

The phrasing of questions in any survey can have a significant
impact on the results. With our questions we hoped to be clear and
brief, so that different annotators do not misinterpret what was being
asked of them.  
% However, before arriving at the eventual way of phrasing
% the questions, 
\ch{In order} to determine the more suitable way to formulate the questions, we performed two separate annotations on a smaller pilot
set of 2100 terms. One, in which we asked if a word is {\it associated} with a certain emotion,
and another independent set of annotations 
where we asked whether a word
{\it evokes} a certain emotion. We found that the annotators agreed
with each other much more in the {\it associated} case than in the
{\it evokes} case.  (Details are in Section~\ref{sec:evoke} ahead.) Therefore all
subsequent annotations were done with {\it associated}. All results,
except those presented in Section~\ref{sec:evoke}, are for the {\it
associated} annotations.

Below is a complete example HIT for the
target word {\it startle}. 
  Note that all questions are multiple-choice
questions, and the Turkers could select exactly one option for each
question.
The survey was approved by the ethics committee at
the National Research Council Canada. \\ \\ \\

% \hline
 \rule{13.5cm}{0.4mm}
\begin{lquote}
% \centerline{\bf Example HIT}
% \ \\
\hspace{-2.5mm} {\bf Title:} Emotions associated with words\\
% {\bf Description:} Specify the emotions evoked by a word by answering multiple choice questions.\\
{\bf Keywords:} emotion, English, sentiment, word association, word meaning\\
% {\bf Time allotted per assignment (HIT):} 5 minutes\\
\indent {\bf Reward per HIT:} \$0.04\\

{\bf Directions:}
\begin{enumerate}[1.]
\item This survey will be used to better understand emotions. Your input is much appreciated.
\item If any of the questions in a HIT are unanswered, then the assignment is no longer useful to us and we will be unable to pay for the assignment.
\item Please return/skip HIT if you do not know the meaning of the word.
\item Attempt HITS only if you are a native speaker of English, or very fluent in English.
\item Certain ``check questions" will be used to make sure the annotation is responsible and reasonable. Assignments that fail these tests will be rejected. If an annotator fails too many of these check questions, then it will be assumed that the annotator is not following instructions 3 and/or 4 above,
and ALL of the annotator's assignments will be rejected.
 \item We hate to reject assignments, but we must at times, to be fair to those who answer the survey with diligence and responsibility. In the past we have approved completed assignments by more than 95\% of the Turkers. 
 If you are unsure about your answers and this is the first time that you are answering
 an emotion survey posted by us, then we recommend that you NOT do a huge number of HITs right away. Once your initial HITS are approved, you gain confidence in your answers and in us. 
\item We will approve HITs about once a week. Expected date all the assignments will be approved: April 14, 2010.
\item Confidentiality notice: Your responses are confidential. Any publications based on these responses will not include your specific responses, but rather aggregate information from many individuals. We will not ask any information that can be used to identify who you are.
\item Word meanings: Some words have more than one meaning, and the different meanings may be associated with different emotions.  For each HIT, Question 1 (Q1) will guide you to the intended meaning. You may encounter multiple HITs for the same target term, but they will correspond to different meanings of
the target word, and they will have different guiding questions.
% \item Note that the order of options in Q13 changes with every HIT.
\end{enumerate}

{\bf Prompt word: {\it startle}}\\

Q1. Which word is closest in meaning (most related) to {\it startle}?
\begin{itemize}
\item {\it automobile} 
\item {\it shake} 
\item {\it honesty} 
\item {\it entertain} 
\end{itemize}

Q2. How positive (good, praising) is the word {\it startle}?
\begin{itemize}
\item {\it startle} is not positive
\item {\it startle} is weakly positive
\item {\it startle} is moderately positive
\item {\it startle} is strongly positive
\end{itemize}

Q3. How negative (bad, criticizing) is the word {\it startle}?
\begin{itemize}
\item {\it startle} is not negative
\item {\it startle} is weakly negative
\item {\it startle} is moderately negative
\item {\it startle} is strongly negative
\end{itemize}

Q4. How much is {\it startle} associated with the emotion joy? (For example, {\it happy} and {\it fun} are strongly associated with joy.)
\begin{itemize}
\item {\it startle} is not associated with joy
\item {\it startle} is weakly associated with joy
\item {\it startle} is moderately associated with joy
\item {\it startle} is strongly associated with joy
\end{itemize}

Q5. How much is {\it startle} associated with the emotion sadness? (For example, {\it failure} and {\it heart-break} are strongly associated with sadness.)
\begin{itemize}
\item {\it startle} is not associated with sadness
\item {\it startle} is weakly associated with sadness
\item {\it startle} is moderately associated with sadness
\item {\it startle} is strongly associated with sadness
\end{itemize}

Q6. How much is {\it startle} associated with the emotion fear? (For example, {\it horror} and {\it scary} are strongly associated with fear.)
\begin{itemize}
 \item Similar choices as in 4 and 5 above
\end{itemize}

Q7. How much is {\it startle} associated with the emotion anger? (For example, {\it rage} and {\it shouting} are strongly associated with anger.)
\begin{itemize}
 \item Similar choices as in 4 and 5 above
\end{itemize}

Q8. How much is {\it startle} associated with the emotion trust? (For example, {\it faith} and {\it integrity} are strongly associated with trust.)
\begin{itemize}
 \item Similar choices as in 4 and 5 above
\end{itemize}

Q9. How much is {\it startle} associated with the emotion disgust? (For example, {\it gross} and {\it cruelty} are strongly associated with disgust.)
\begin{itemize}
 \item Similar choices as in 4 and 5 above
\end{itemize}

Q10. How much is {\it startle} associated with the emotion surprise? (For example, {\it startle} and {\it sudden} are strongly associated with surprise.)
\begin{itemize}
 \item Similar choices as in 4 and 5 above
\end{itemize}

Q11. How much is {\it startle} associated with the emotion anticipation? (For example, {\it expect} and {\it eager} are strongly associated with anticipation.)
\begin{itemize}
 \item Similar choices as in 4 and 5 above
\end{itemize}

Q12. Is {\it startle} an emotion? (For example: {\it love} is an emotion; {\it shark} is associated with fear (an emotion), but {\it shark} is not an emotion.)
\begin{itemize}
 \item No, {\it startle} is not an emotion
 \item Yes, {\it startle} is an emotion
\end{itemize}
\rule{13.5cm}{0.4mm}
% 13. What colour is associated with {\it startle}?\\
% \begin{minipage}[t]{0.2cm}
% \ \\
% \end{minipage}
% \begin{minipage}[t]{3cm}
%  \begin{itemize}
%  \item white
%  \item black
%  \item red
%  \item green
% \end{itemize}
% \end{minipage}
% \begin{minipage}[t]{3cm}
%  \begin{itemize}
%  \item yellow
%  \item blue
%  \item brown
%  \item pink\\
% \end{itemize}
% \end{minipage}
% \begin{minipage}[t]{3cm}
%  \begin{itemize}
%  \item purple
%  \item orange
%  \item grey\\
% \end{itemize}
% \end{minipage}
% For each term, the colour options in question 13 were presented in random order.
% Before going live, 
\end{lquote}

% Why make 8 questions instead of 1.\\
% Why 4 levels of intensity.

% \section{Annotation analysis}
\section{Annotation Statistics and Post-Processing}
We conducted annotations in two batches, starting first with a pilot
set of about 2100 terms, which was annotated in about a week.  The
second batch of about 8000 terms (HITs) was annotated in about two
weeks. Notice that the amount of time taken is not linearly
proportional to the number of HITs. We \ch{speculate} that as one builds a history of tasks
and payment, more Turkers do subsequent tasks. Also, if there
are a large number of HITs, then \ch{probably} more people find it worth the effort
to understand and become comfortable at doing the task.  Each HIT had
a compensation of \$0.04 (4 cents) and the Turkers spent about a
minute on average to answer the questions in a HIT.  This resulted in
an hourly pay of about \$2.40.

% the data, and expectedly the number of HITs attempted by them followed a Zipfian distribution:

Once the assignments were collected, we used automatic scripts to
validate the annotations.  Some assignments were discarded because
they failed certain tests (described below). A subset of the discarded
assignments were officially {\ebf rejected} (the Turkers were not paid
for these assignments) because instructions were not followed.  About
2,666 of the 50,850 (10,170 $\times$ 5) assignments included at least
one unanswered question.  These assignments were discarded and
rejected.  Even though distractors for Q1 were chosen at random, every
now and then a distractor may come too close to the meaning of the
target term, resulting in a bad word choice question.  For 1045 terms,
three or more annotators gave an answer different from the one
generated automatically from the thesaurus. These questions were marked as
bad questions and discarded. All corresponding assignments (5,225 in
total) were discarded.  Turkers were paid in full for these
assignments regardless of their answer to Q1.

More than 95\% of the remaining assignments had the correct answer for
the word choice question. This was a welcome result, showing that
most of the annotations were done in an \ch{appropriate} manner.  We
discarded all assignments that had the wrong answer for the word
choice question.  If an annotator obtained an overall score that is
less than 66.67\% on the word choice questions (that is, got more than
one out of three wrong), then we assumed that, contrary to
instructions, the annotator attempted to answer HITs for words that were unfamiliar.
We discarded and rejected {\it all} assignments by such
annotators  (not merely the assignments for which they got the word
choice question wrong).

% For many of the terms, the emotion questions (Q5 through Q11)  may not
% have a ``right answer". (Even though many may have a clear wrong
% answer. For example, {\it morose} is definitely not associated with
% joy.) Therefore, 
For each of the annotators, we calculated the maximum
likelihood probability with which the annotator agrees with the
majority on the emotion questions.  We calculated the mean of these
probabilities and the standard deviation.  Consistent with standard
practices in identifying outliers, we discarded annotations by Turkers
who were more than two standard deviations away from the mean
(annotations by 111 Turkers).

% Finally, there remained three or more valid assignments for 8,883  of the 10,170 target terms.  
After this post-processing, 8,883 of the initial 10,170 terms remained, each with three or more valid assignments.
We will refer to this set of assignments as
the {\ebf master set}.  We created the word--emotion association
lexicon from this master set, containing 38,726 assignments from about
2,216 Turkers who attempted 1 to 2,000 assignments each.  About 300 of
them provided 20 or more assignments each (more than 33,000
assignments in all).  The master set has, on average, about 4.4
assignments for each of the 8,883 target terms.  (See Table~\ref{tab:emotion terms} for more
details.) The total cost of the annotation was about US\$2,100. This
includes fees that Amazon charges (about 13\% of the amount paid to
the Turkers) as well as the cost for the dual annotation of the pilot
set with both {\it evokes} and {\it associated}.  \footnote{We will upload
HITs of discarded assignments on Mechanical Turk for another
round of annotations.}

\begin{table}[t]
\caption{Break down of target terms into various categories. Initial refers to terms chosen for annotation. Master refers to terms for which three or more valid assignments were obtained using Mechanical Turk.
MQ stands for Macquarie Thesaurus, GI for General Inquirer, and WAL for WordNet Affect Lexicon.}
\centering
% \resizebox{0.5\textwidth}{!}{
\begin{tabular}{lrrr}
\hline
						
  						&\multicolumn{2}{c}{\bf \# of terms} 	&{\bf Annotations}\\
	{\bf EmoLex} 		&{\bf Initial}   	&{\bf Master}	&{\bf per word} \\
\hline
	\multicolumn{4}{l}{\bf EmoLex-Uni:}  \\
	\multicolumn{3}{l}{Unigrams from Macquarie Thesaurus}  \\
	$\;\;\;\;$ adjectives         &200                &190            &4.4 \\
	$\;\;\;\;$ adverbs          &200                &187            &4.5 \\
	$\;\;\;\;$ nouns           &200                &178            &4.5 \\
	$\;\;\;\;$ verbs            &200                &195            &4.4 \\

	\multicolumn{4}{l}{\bf EmoLex-Bi:} \rule{0pt}{12pt} \\
	\multicolumn{3}{l}{Bigrams from Macquarie Thesaurus}  \\
	$\;\;\;\;$ adjectives       &200                &162            &4.4 \\
	$\;\;\;\;$ adverbs         &187                &171            &4.3 \\
	$\;\;\;\;$ nouns           &200                &185            &4.5 \\
	$\;\;\;\;$ verbs           &200                &178            &4.4 \\

	\multicolumn{4}{l}{\bf EmoLex-GI:} \rule{0pt}{12pt}\\
	\multicolumn{4}{l}{Terms from General Inquirer}  \\
	$\;\;\;\;$ negative terms          &2119                &1837            &4.4 \\
	$\;\;\;\;$ neutral terms           &4226                &3653            &4.4 \\
	$\;\;\;\;$ positive terms          &1787                &1541            &4.4 \\

	\multicolumn{4}{l}{\bf EmoLex-WAL:} \rule{0pt}{12pt} \\
	\multicolumn{4}{l}{Terms from WordNet Affect Lexicon} \\
	$\;\;\;\;$ anger terms          &165            &160            &4.5 \\
	$\;\;\;\;$ disgust terms        &37             &34             &4.4 \\
	$\;\;\;\;$ fear terms           &100            &89             &4.4 \\
	$\;\;\;\;$ joy terms            &165            &149            &4.5 \\
	$\;\;\;\;$ sadness terms        &120            &112            &4.5 \\
	$\;\;\;\;$ surprise terms       &53             &51             &4.4 \\
	{\bf Union} \rule{0pt}{10pt}											&{\bf 10170}				&{\bf 8883}			&{\bf 4.45} \\

% 	$\;\;\;\;$ {\bf all}   				 	& &	& && \\
% {\bf ALL}   				 & &	& && \\
\hline
\\
\end{tabular}
% }
\label{tab:emotion terms}
\end{table}

\begin{table}[t!]
\caption{Percentage of terms with majority class of no, weak, moderate, and strong emotion.}
\centering
% \resizebox{0.5\textwidth}{!}{
% {\small
\begin{tabular}{l rrrr}
\hline
 
        &\multicolumn{4}{c}{\bf Intensity} \\
		        Emotion &no      &weak   &moderate   &strong \\
				\hline
%				anger        &78.8       &9.4        &6.2        &5.4        \\
%				anticipation      &71.4       &13.6       &9.4        &5.3        \\
%				disgust      &82.6       &8.8        &4.9        &3.5        \\
%				fear         &76.5       &11.3       &7.3        &4.7        \\
%				joy      &72.6       &9.6        &10.0       &7.6        \\
%				sadness      &76.0       &12.4       &5.8        &5.6        \\
%				surprise         &84.8       &7.9        &4.1        &3.0        \\
%				trust        &73.3       &12.0       &9.8        &4.7        \\

%				{\bf micro-average}        &{\bf 77.0}     &{\bf 10.6}       &{\bf 7.2}       &{\bf 5.0}        \\
%				{\bf any emotion}        &{\bf 17.9}     &{\bf 23.4}     &{\bf 28.3}     &{\bf 30.1}     \\

				anger        &81.6       &8.5        &5.1        &4.5        \\
				anticipation &84.2       &8.9        &4.2        &2.4        \\
				disgust      &84.6       &8.3        &3.8        &3.1        \\
				fear         &79.6       &10.3       &5.6        &4.3        \\
				joy      	 &79.5       &8.9        &6.4        &5.0        \\
				sadness      &80.9       &10.0       &4.8        &4.2        \\
				surprise     &89.5       &6.6        &2.2        &1.4        \\
				trust        &81.9       &7.9        &5.9        &4.1        \\

				{\bf micro-average}      &{\bf 82.7}     &{\bf 8.7}      &{\bf 4.8}     &{\bf 3.6}        \\ 
				{\bf any emotion}        &{\bf 35.6}     &{\bf 21.2}     &{\bf 20.5}     &{\bf 22.5}     \\
\hline
\\
\end{tabular}
% }
\label{tab:emo 1}
\end{table}

\begin{table*}[t!]
\caption{Percentage of terms, in each target set, that are emotive. Highest individual emotion scores for EmoLex-WAL  
are shown in bold. 
The \ch{last column}, {\it any}, shows the percentage of terms associated
with at least one of the eight emotions.
Observe that WAL fear terms are marked most as associate with fear, joy terms as associated with joy, and so on.}
\centering
 \resizebox{\textwidth}{!}{
% {\small
\begin{tabular}{l rrr rrr rrr r}
\hline
  %						&\multicolumn{9}{c}{\bf \% of evocative terms}\\
	 					&{\bf anger} 	&{\bf anticipn.} 	&{\bf disgust}				&{\bf fear}  	&{\bf joy} &{\bf sadness} 	&{\bf surprise} 	&{\bf trust}  &{\bf any}\\
\hline
% FOR only google terms
% {\bf EmoLex} 			&10.5	&21.8	&6.4	&11.6	&22.0	&9.8	&7.1	&23.0	&59.8\\
% {\bf EmoLex} 			&13.1	&11.8	&10.1	&13.8	&16.1	&12.1	&6.2	&15.6	&54.4\\
{\bf EmoLex} 			&13	&12	&10	&14	&16	&12	&6	&16	&54\\

	\multicolumn{5}{l}{\bf EmoLex-Uni:} \rule{0pt}{12pt} & & & & &\\
\multicolumn{4}{l}{Unigrams from Macquarie Thesaurus} \\
$\;\;\;\;$ adjectives      &14      &14    &10     &13 &29     &14    &10    &15   &68\\ 
$\;\;\;\;$ adverb          &13      &20    &8  &10 &23     &11    &7    &23    &67\\
$\;\;\;\;$ noun        	&7   &18    &3  &7  &16     &6    &3    &24 &46\\
$\;\;\;\;$ verb       	&11      &21    &5  &16 &14     &11    &7    &15    &52\\

	\multicolumn{5}{l}{\bf EmoLex-Bi:} \rule{0pt}{12pt} & & & & &\\
\multicolumn{4}{l}{Bigrams from Macquarie Thesaurus} \\
$\;\;\;\;$ adjectives       &12      &25    &8  &14 &30     &15    &8    &16    &66\\
$\;\;\;\;$ adverbs         &6   &23    &1  &7  &19     &3    &9    &29 &54\\
$\;\;\;\;$ nouns       	&9   &23    &6  &14 &20     &9    &7    &29 &58\\
$\;\;\;\;$ verbs      	&8   &25    &5  &7  &21     &6    &3    &27 &60\\

	\multicolumn{5}{l}{\bf EmoLex-GI:} \rule{0pt}{12pt} & & & & &\\
\multicolumn{4}{l}{Terms from General Inquirer}  \\
$\;\;\;\;$ negative terms          &36      &4 &29     &34 &0     &33    &8    &2  &67\\
$\;\;\;\;$ neutral terms           &4   &11    &3  &8  &10     &4    &5    &13 &36\\
$\;\;\;\;$ positive terms          &1   &13    &0  &2  &40     &1    &4    &33 &62\\

	\multicolumn{5}{l}{\bf EmoLex-WAL:} \rule{0pt}{12pt} & & & & &\\
	\multicolumn{4}{l}{Terms from WordNet Affect Lexicon} \\
$\;\;\;\;$ anger terms          &{\bf 83}      &1 &53     &18 &0     &16    &0    &0  &90\\
$\;\;\;\;$ disgust terms         &44      &0 &{\bf 94}     &14 &0     &2    &0    &0   &94\\
$\;\;\;\;$ fear terms       		&17      &17    &19     &{\bf 74} &1     &20    &15    &3 &89\\ 
$\;\;\;\;$ joy terms         	&2   &14    &0  &2  &{\bf 78}     &2    &7    &28 &91\\
$\;\;\;\;$ sadness terms       &9   &0 &13     &13 &0     &{\bf 94}    &0    &0  &96\\
$\;\;\;\;$ surprise terms     &2   &6 &4  &8  &42     &6    &66    &6 &{\bf 88}\\

% {\bf ALL}   				 	& &	& & & & & & &\\
\hline
\end{tabular}
 }
\label{tab:emo break}
\end{table*}

% colour        &22.8       &18.5       &13.4       &11.9       &10.0       &6.4        &6.3        &5.3        &2.1        &1.5		&1.3        \\

% \caption{Percent of 2081 target terms that are evocative.}

\section{Analysis of Emotion Annotations}
\label{sec: emo analysis}

The different emotion annotations for a target term were consolidated
by determining the {\ebf majority class} of emotion intensities. For a
given term--emotion pair, the majority class is that intensity level
that is chosen most often by the Turkers to represent the degree of
emotion evoked by the word.  Ties are broken by choosing the stronger
intensity level.  Table~\ref{tab:emo 1} lists the percentage of 8,883
target terms assigned a majority class of no, weak, moderate, and
strong emotion.  For example, it tells us that 5\% of the target terms
strongly evoke joy.  The table also presents averages of the
numbers in each column (micro-averages).  The last row lists the
percentage of target terms that evoke some emotion (any of the eight) at
the various intensity levels. We calculated this using the intensity
level of the strongest emotion expressed by each target.  Observe that
22.5\% of the target terms strongly evoke at least one of the eight
basic emotions.

Even though we asked Turkers to annotate emotions at four levels of
intensity, practical NLP applications often require only two
levels---associated with a given emotion (we will refer to these terms as being {\ebf emotive})
or not associated with the emotion (we will refer to these terms as being {\ebf non-emotive}).  
For each target term--emotion
pair, we convert the four-level annotations into two-level annotations
by placing all no- and weak-intensity assignments in the non-emotive
bin, all moderate- and strong-intensity assignments in the emotive
bin, and then choosing the bin with the majority assignments.  Table~\ref{tab:emo break} shows the \ch{percentage} of terms 
associated with the different emotions.
The \ch{last column}, {\it any}, shows the percentage of terms associated
with at least one of the eight emotions.

Analysis of Q12 revealed that 9.3\% of the 8,883 target terms
(826 terms) were considered not merely to be associated with certain emotions, but
also to refer directly to emotions.

\subsection{Discussion}
Table~\ref{tab:emo break} shows that a sizable percentage of
nouns, verbs, adjectives, and adverbs are emotive.  
Trust (16\%), and joy (16\%) are the most common emotions associated with terms.  
\ch{Among} the four parts of speech, adjectives (68\%) and
adverbs (67\%) are most often associated with emotions and this is not
surprising considering that they are used to qualify nouns and verbs,
respectively.  
Nouns are more commonly associated with trust (16\%),
whereas adjectives are more commonly associated with joy (29\%).

The EmoLex-WAL rows are particularly
interesting because they serve to determine how much the Turker
annotations match annotations in the Wordnet Affect Lexicon (WAL).
The most common Turker-determined emotion for each of these rows is
marked in bold. Observe that WAL anger terms are mostly marked as
associated with anger, joy terms as associated with joy, and so on.  Here is the
complete list of terms that are marked as anger terms in WAL, but were
not marked as anger terms by the Turkers: {\it baffled, exacerbate,
gravel, pesky, and pestering}. One can see that indeed many of these
terms are not truly associated with anger.  We also observed that the Turkers
marked some terms as being associated with both anger and joy.  The
complete list includes: {\it adjourn, credit card, find out, gloat,
spontaneously, and surprised}. One can see how many of these words are
indeed associated with both anger and joy.
The EmoLex-WAL rows also indicate which
emotions tend to be jointly associated to a term.  Observe that anger
terms tend also to be associated with disgust.  Similarly, many joy terms
are also associated with trust. The surprise terms in WAL are largely also
associated with joy.

The EmoLex-GI rows rightly show that words
marked as negative in the General Inquirer are mostly associated with
negative emotions (anger, fear, disgust, and sadness). Observe that
the percentages for trust and joy are much lower. On the other hand,
positive words are associated with anticipation, joy, and trust.  

% It is interesting to see how a good number of surprise evoking words are
% positive, and yet a good number of them  % are negative.

\subsection{Agreement}

In order to analyze how often the annotators agreed with
each other, for each term--emotion pair, we calculated the percentage
of times the majority class has size 5 (all Turkers agree), size 4
(all but one agree), size 3, and size 2.  Table~\ref{tab:emo agree 1}
presents these agreement values.  Observe that for almost 60\% of the
terms, at least four annotators agree with each other (see bottom right corner of Table~\ref{tab:emo agree 1}).  Since many NLP
systems may rely only on two intensity values (emotive or
non-emotive), we also calculate agreement at that level
(Table~\ref{tab:emo agree 2}).  For more than 60\% of the
terms, all five annotators agree with each other, and for almost
85\% of the terms, at least four annotators agree (see bottom right corner of Table~\ref{tab:emo agree 2}).  
\ch{These} agreements are despite the somewhat subjective nature of word--emotion associations, 
and despite the absence of any control over the educational background of the
annotators. \ch{We provide} agreement values along with each of the term–emotion pairs
so that downstream applications can selectively use the lexicon.

\begin{table}[t]
\caption{Agreement at four intensity levels of emotion (no, weak, moderate, and strong): Percentage of terms for which the majority class size was 2, 3, 4, and 5.
Note that, given five annotators and four levels, the majority class size must be between two and five.}
\centering
% \resizebox{0.5\textwidth}{!}{
% {\small
\begin{tabular}{l rrrr rr}
\hline
 
				&\multicolumn{6}{c}{\bf Majority class size} \\
				{\bf Emotion} &$=$ two     &$=$ three      &$=$ four   &$=$ five &$\geq$ three  &$\geq$ four\\
				\hline
				anger        &13.7        &21.7        &25.7        &38.7        &86.1 &64.4\\
				anticipation &19.2        &31.7        &28.3        &20.7        &80.7 &49.0\\
				disgust      &13.8        &20.7        &23.8        &41.5        &86.0 &65.3\\
				fear         &16.7        &27.7        &25.6        &29.9        &83.2 &55.5\\
				joy      	 &16.1        &24.3        &21.9        &37.5        &83.7 &59.4\\
				sadness      &14.3        &23.8        &25.9        &35.7        &85.4 &61.6\\
				surprise     &11.8        &25.3        &32.2        &30.6        &88.1 &62.8\\
				trust        &18.8        &27.4        &27.7        &25.9        &81.0 &53.6\\
% 				{\bf All}        &{\bf 15.6}     &{\bf 25.3}     &{\bf 26.4}     &{\bf 32.5}     \\
				{\bf micro-average}      &{\bf 15.6}        &{\bf 25.3}       &{\bf 26.4}      &{\bf 32.6}      &{\bf 84.3} &{\bf 59.0}\\
\hline
\\
\end{tabular}
% }
\label{tab:emo agree 1}
\end{table}

\begin{table}
\caption{Agreement at two intensity levels of emotion (emotive and non-emotive): Percentage of terms for which the majority class size was 3, 4, and 5.
Note that, given five annotators and two levels, the majority class size must be between three and five.}
\centering
% \resizebox{0.5\textwidth}{!}{
% {\small
\begin{tabular}{l rrrr}
\hline
 
			        &\multicolumn{4}{c}{\bf Majority class size} \\
					        {\bf Emotion} &$=$ three   &$=$ four   &$=$ five &$\geq$ four\\
							\hline
							anger   		&13.2   &19.4   &67.2    &86.6\\
							anticipation 	&18.8   &32.6   &48.4    &81.0\\
							disgust 		&13.4   &18.4   &68.1    &86.5\\
							fear    		&15.3   &24.8   &59.7    &84.5\\
							joy 			&16.2   &22.6   &61.0    &83.6\\
							 % negative    &11.5   &22.3   &66.1    \\
							 % positive    &24.2   &26.3   &49.3    \\
							sadness 		&12.8   &20.2   &66.9    &87.1\\
							surprise    	&10.9   &22.8   &66.2    &89.0\\
							trust   		&20.3   &28.8   &50.7    &79.5\\
							{\bf micro-average}      &{\bf 15.1}        &{\bf 23.7}      &{\bf 61.0}      &{\bf 84.7}\\

\hline
\\
\end{tabular}
% }
\label{tab:emo agree 2}
\end{table}

Cohen's $\kappa$ \cite[]{Cohen60} is a widely used measure for inter-annotator agreement.
It corrects observed agreement for chance agreement by using the distribution of classes
chosen by each of the annotators. 
\ch{However,} it is appropriate only when the same judges annotate all the instances \cite[]{Fleiss71}.
In Mechanical Turk, annotators are given the freedom
to annotate as many terms as they wish, and many annotate only a small number of terms (the long tail of the zipfian distribution).
Thus the judges do not annotate all of the instances, and further, one cannot reliably estimate the
distribution of classes chosen by each judge when they annotate only a small number of instances.
Scott's $\Pi$ \cite[]{Scott55} calculates
chance agreement by determining the distribution each of the categories (regardless of who the annotator is).
This is more appropriate for our data, but it applies only to scenarios with exactly two annotators.
\cite{Fleiss71} proposed a generalization of Scott's $\Pi$ for when there are more than two annotators,
which he called $\kappa$ even though Fleiss's $\kappa$ is more like Scott's $\Pi$ than Cohen's $\kappa$.
All subsequent mentions of $\kappa$ in this paper will refer to Fleiss's $\kappa$ unless explicitly stated otherwise.
\cite{LandisK77} provided Table~\ref{tab:scotts pie segments} to interpret the $\kappa$ values.
Table~\ref{tab:emo scott pie} lists the $\kappa$ values
for the Mechanical Turk emotion annotations.

\begin{table}
\caption{Segments of Fleiss $\kappa$ values and their interpretations (Landis and Koch, 1977).}
\centering
\begin{tabular}{l l}
\hline

                            {\bf Fleiss's} ${\bf \kappa}$ &{\bf Interpretation}   \\
                            \hline
	 $<$ 0	&poor agreement\\
0.00 -- 0.20	&slight agreement\\
0.21 -– 0.40	&fair agreement\\
0.41 -– 0.60	&moderate agreement\\
0.61 -– 0.80	&substantial agreement\\
0.81 -– 1.00	&almost perfect agreement\\
\hline
\\
\end{tabular}
\label{tab:scotts pie segments}
\end{table}
\begin{table}
\caption{Agreement at two intensity levels of emotion (emotive and non-emotive): Fleiss's $\kappa$, and its interpretation.}
\centering
\begin{tabular}{l rl}
\hline

                            {\bf Emotion} &{\bf Fleiss's} ${\bf \kappa}$   &{\bf Interpretation} \\
                            \hline
                            anger           &0.39   &fair agreement    \\
                            anticipation    &0.14   &slight agreement    \\
                            disgust         &0.31   &fair agreement    \\
                            fear            &0.32   &fair agreement    \\
                            joy             &0.36   &fair agreement    \\
                            sadness         &0.39   &fair agreement  \\
                            surprise        &0.18   &slight agreement    \\
                            trust           &0.24   &fair agreement    \\
                            {\bf micro-average}       &{\bf 0.29}      &{\bf fair agreement}      \\

\hline
\\
\end{tabular}
\label{tab:emo scott pie}
\end{table}

The $\kappa$ values show that for six of the eight emotions the Turkers have fair agreement, and for
anticipation and trust there is only slight agreement.
The $\kappa$ values for anger and sadness are the highest.
The average $\kappa$ value for the eight emotions is 0.29, and it implies fair agreement.
% \ch{Unlike} certain other annotation tasks,
% there is no correct answer to many of the word--emotion questions
% given to the annotators. Indeed, people will agree more with each other
% regarding the emotion associations for some words and agree less for other words.
Below are some reasons why agreement values are much lower than certain other tasks,
for example, part of speech tagging:
\begin{itemize}
\item The target word is presented out of context. We expect higher agreement if
we provided words in particular contexts, but words can occur in innumerable contexts,
and annotating too many instances of the same word is costly.
%, and that will entail an explosion of instances
% to be annotated. 
By providing the word choice question, we bias the Turker towards a particular sense
of the target word, and aim to obtain the prior probability of the word sense's emotion association.
\item Words are associated with emotions to different degrees, and there are no clear classes corresponding to different levels of association.
Since we ask people to place term-emotion associations in four specific bins, more people disagree for term--emotion pairs
whose degree of association is closer to the boundaries, than for other term--emotion pairs.
\item \cite{Holsti69}, \cite{BrennanP81}, \cite{PerreaultL89}, and others consider 
the $\kappa$ values (both Fleiss's and Cohen's) to be conservative, especially 
when one category is much more prevalent than the other.
In our data, the ``not associated with emotion" category is much more prevalent than the ``associated with emotion" category,
so these $\kappa$ values might be underestimates of the true agreement.
\end{itemize}
\noindent Nonetheless, as mentioned earlier,  when using the lexicon in downstream applications, one may employ suitable strategies 
such as choosing instances that have high agreement scores, averaging information from many words, and using contextual information
in addition to information obtained form the lexicon.

\subsection{Evokes versus Associated}
\label{sec:evoke}
As alluded to earlier, we performed two separate sets of annotations on
the pilot set: one where we asked if a word {\it evokes} a certain
emotion, and another where we asked if a word is {\it associated} with
a certain emotion. Table~\ref{tab:evoke vs assoc} lists the the
percentage of times all five annotators agreed with each other on the
classification of a term as emotive,
for the two scenarios.
Observe that the agreement numbers are markedly higher with {\it
associated} than with {\it evokes} for anger, anticipation, joy, and
surprise. In case of fear and sadness, the agreement is only slightly
better with {\it evokes}, whereas for trust and disgust the agreement
is slightly better with {\it associated}.  Overall, {\it associated}
leads to an increase in agreement by more than 5 percentage points over
{\it evokes}. Therefore all subsequent annotations were performed
with {\it associated} only. (All results shown in this paper, except
for those in Table~\ref{tab:evoke vs assoc}, are for {\it
associated}.) 

\begin{table}
\caption{{\it Evokes} versus {\it associated}: Agreement at two intensity levels of emotion (emotive and non-emotive).
Percentage of terms in the pilot set for which the majority class size was 5.}
\centering
% \resizebox{0.5\textwidth}{!}{
% {\small
\begin{tabular}{l rr}
\hline
 
			        &\multicolumn{2}{c}{\bf Majority class size five} \\
					        {\bf Emotion} &evokes   &associated \\
							\hline
							anger     		&61.6   &{\bf 68.2}    \\
							anticipation 	&34.8   &{\bf 49.6}    \\
							disgust 		&65.4   &{\bf 66.4}    \\
							fear    		&{\bf 62.0}   &59.4    \\
							joy 			&54.6   &{\bf 62.3}    \\
							 % negative    	&67.3   &68.2    \\
							 % positive    	&49.0   &57.7    \\
							sadness 		&{\bf 66.7}   &65.3    \\
							surprise    	&54.0   &{\bf 67.3}    \\
							trust   		&47.3   &{\bf 49.8}    \\
							{\bf micro-average}       &55.8      &{\bf 61.0}      \\

\hline
\\
\end{tabular}
% }
\label{tab:evoke vs assoc}
\end{table}

We speculate that to answer which emotions are {\it evoked} by a term,
people sometimes bring in their own varied personal experiences, and
so we see relatively more disagreement than when we ask what emotions
are {\it associated} with a term. In the latter case, people may be
answering what is more widely accepted rather than their own personal
perspective.  Further investigation on the differences between {\it
evoke} and {\it associated}, and why there is a marked difference in
agreements for some emotions and not so much for others, is left as
future work.

\section{Analysis of Polarity Annotations}

\begin{table}[t]
\caption{Percentage of terms given majority class of no, weak, moderate, and strong polarity.}
\centering
% \resizebox{0.5\textwidth}{!}{
% {\small
\begin{tabular}{l rrrr}
\hline
        &\multicolumn{4}{c}{\bf Intensity} \\
        Polarity &no      &weak   &moderate   &strong \\
\hline

				negative         &64.3       &9.1        &10.8       &15.6       \\
				positive         &61.9       &9.8        &13.7       &14.4       \\

				 % negative         &60.8       &10.8       &12.3       &16.0       \\
				 % positive         &48.3       &11.7       &20.7       &19.0       \\
				 % {\bf micro-average}       &{\bf 54.6}        &{\bf 11.3}        &{\bf 16.5}     &{\bf 17.5}     \\
		 		 % {\bf any polarity}     &{\bf 14.7}     &{\bf 17.4}     &{\bf 32.7}     &{\bf 35.0}     \\
				 {\bf polarity average}      &{\bf 63.1}     &{\bf 9.5}     &{\bf 12.3}        &{\bf 15.0}     \\ 
				 {\bf either polarity}    &{\bf 29.9}     &{\bf 15.4}     &{\bf 24.3}     &{\bf 30.1}     \\ 
\hline
\\
\end{tabular}
% }
\label{tab:pol 1}
\end{table}

\begin{table}[t!]
\caption{Percentage of terms, in each target set, that are evaluative. The highest scores for EmoLex-GI
positives and negatives are shown bold. Observe that the positive GI terms are marked mostly as positively evaluative
and the negative terms are marked mostly as negatively evaluative.}
\centering
% \resizebox{.5\textwidth}{!}{
% {\small
\begin{tabular}{l rrr}
\hline
%  						&\multicolumn{2}{c}{\bf \% of terms}\\
	  			&{\bf negative} 			&{\bf positive} 	&{\bf either}\\
\hline
{\bf EmoLex} 						&30	&35	&65\\

% & & &\\
	\multicolumn{4}{l}{\bf EmoLex-Uni:} \rule{0pt}{12pt}\\
	\multicolumn{4}{l}{Unigrams from Macquarie Thesaurus}  \\
$\;\;\;\;$ adjectives                   &32             &48         &79\\
$\;\;\;\;$ adverbs                      &26             &55         &80\\
$\;\;\;\;$ nouns                   		&8          	&39         &46\\ 
$\;\;\;\;$ verbs                   		&26             &37         &63\\ 

	\multicolumn{4}{l}{\bf EmoLex-Bi:} \rule{0pt}{12pt}\\
\multicolumn{4}{l}{Bigrams from Macquarie Thesaurus} \\
$\;\;\;\;$ adjectives                   &30             &47         &77\\
$\;\;\;\;$ adverbs                      &11             &42         &52\\
$\;\;\;\;$ nouns                    	&14             &45         &57\\
$\;\;\;\;$ verbs                    	&14             &48         &60\\

	\multicolumn{4}{l}{\bf EmoLex-GI:} \rule{0pt}{12pt}\\
\multicolumn{4}{l}{Terms from General Inquirer}  \\
$\;\;\;\;$ negative terms              &{\bf 83}       &1          &85\\
$\;\;\;\;$ neutral terms               &12             &30         &41\\
$\;\;\;\;$ positive terms           &2          	&82         &{\bf 84}\\

	\multicolumn{4}{l}{\bf EmoLex-WAL:} \rule{0pt}{12pt} \\
	\multicolumn{4}{l}{Terms from WordNet Affect Lexicon} \\
$\;\;\;\;$ anger terms                  &96             &1          &97\\   
$\;\;\;\;$ disgust terms                 &97             &0          &97\\
$\;\;\;\;$ fear terms                    &85             &1          &86\\
$\;\;\;\;$ joy terms                    &4          &93         &97\\
$\;\;\;\;$ sadness terms                 &91             &4          &95\\
$\;\;\;\;$ surprise terms                &26             &57         &80\\

%  	$\;\;\;\;$ {\bf all}   				 	& &	\\
%  {\bf ALL}   				 	& &	 \\
\hline
\\
\end{tabular}
% }

\label{tab:pol break}
\end{table}

We consolidate the polarity annotations in the
same manner as for emotion annotations.  Table~\ref{tab:pol 1} lists
the percentage of 8,883 target terms assigned a majority class of no,
weak, moderate, and strong polarity.  It states, for
example, that 15.6\% of the target terms are strongly negative.  The
last row in the table lists the percentage of target terms that have some
polarity (positive or negative) at the various intensity
levels.  Observe that 30.1\% of the target terms are either strongly
positive or strongly negative.

Just as in the case for emotions, practical NLP applications often
require only two levels of polarity---having particular
polarity ({\ebf evaluative}) or not ({\ebf non-evaluative}).
For each target term--emotion pair, we convert the four-level semantic
orientation annotations into two-level ones, just as we did for the
emotions.  Table~\ref{tab:pol break} shows how many terms overall and
within each category are positively and negatively evaluative.

\subsection{Discussion}

Observe in Table~\ref{tab:pol break} that, across the board,
a sizable number of terms are evaluative with respect to some semantic
orientation.  Unigram nouns have a markedly lower proportion of
negative terms, and a much higher proportion of positive terms.  It
may be argued that the default polarity of noun concepts
is neutral or positive, and that usually it takes a negative
adjective to make the phrase negative.

The EmoLex-GI rows in the two tables show
that words marked as having a negative polarity in the
General Inquirer are mostly marked as negative by the Turkers. And
similarly, the positives in GI are annotated as positive.  
% Again, this is confirmation that the quality of annotation obtained is high.
Observe that the Turkers mark 12\% of the GI neutral terms as negative
and 30\% of the GI neutral terms as positive. This may be because the
boundary between positive and neutral terms is more fuzzy than between
negative and neutral terms.  The EmoLex-WAL
rows show that anger, disgust, fear, and sadness terms tend not to
have a positive polarity and are mostly negative.  In
contrast, and expectedly, the joy terms are positive.  The surprise
terms are more than twice as likely to be positive than negative.

\subsection{Agreement}
% \noindent In order to analyze how often the annotators agreed with
% each other, 
 For each term--polarity pair, we calculated the percentage
of times the majority class has size 5 (all Turkers agree), size 4
(all but one agree), size 3, and size 2.  Table~\ref{tab:pol agree 1}
presents these agreement values.  For more than 50\% of
the terms, at least four annotators agree with each other (see bottom right corner of Table~\ref{tab:pol agree 1}).
Table~\ref{tab:pol agree 2} gives agreement values at the
two-intensity level.  For more than 55\% of the terms,
all five annotators agree with each other, and for more than 80\% of
the terms, at least four annotators agree (see bottom right corner of Table~\ref{tab:pol agree 2}).
Table~\ref{tab:so scotts pie} lists the Fleiss $\kappa$ values
for the polarity annotations. They are interpreted based on the
segments provided by \cite{LandisK77} (listed earlier in Table~\ref{tab:scotts pie segments}).
Observe that annotations for negative polarity have markedly higher
agreement than annotations for positive polarity. This too may be because 
of the somewhat more fuzzy boundary between positive and neutral, than between negative and neutral.

\begin{table}
\caption{Agreement at four intensity levels of polarity  (no, weak, moderate, and strong): Percentage of terms for which the majority class size was 2, 3, 4, and 5.}
\centering
% \resizebox{0.5\textwidth}{!}{
% {\small
\begin{tabular}{l rrrr rr}
\hline
 
        &\multicolumn{6}{c}{\bf Majority class size} \\
				{\bf Polarity} &$=$ two     &$=$ three      &$=$ four   &$=$ five &$\geq$ three &$\geq$ four\\
				\hline
				negative         &12.8        &27.3        &27.2        &32.5        &87.0 &59.7\\
				positive         &23.5        &28.5        &18.0        &29.8        &76.3 &47.8\\
				{\bf micro-average}       &{\bf 18.2}     &{\bf 27.9}     &{\bf 22.6}     &{\bf 31.2}     &{\bf 81.7} &{\bf 53.8}\\

\hline
\\
\end{tabular}
% }
\label{tab:pol agree 1}
\end{table}

\begin{table}
\caption{Agreement at two intensity levels of polarity (evaluative and non-evaluative): Percentage of terms for which the majority class size was 3, 4, and 5.}
\centering
% \resizebox{0.5\textwidth}{!}{
% {\small
\begin{tabular}{l rrrr}
\hline
 
			        &\multicolumn{3}{c}{\bf Majority class size} \\
					        {\bf Polarity}&three   &four   &five &$\geq$ four\\
							\hline
							negative    &11.5   &22.3   &66.1    &88.4\\
							positive    &24.2   &26.3   &49.3    &75.6\\
							{\bf micro-average}       &{\bf 17.9}        &{\bf 24.3}        &{\bf 57.7}    &{\bf 82.0}\\

\hline
\\
\end{tabular}
% }
\label{tab:pol agree 2}
\end{table}

\begin{table}
\caption{Agreement at two intensity levels of polarity (evaluative and non-evaluative): Fleiss's $\kappa$, and its interpretation.}
\centering
\begin{tabular}{l rl}
\hline

                            {\bf Polarity} &{\bf Fleiss's $\kappa$}   &{\bf Interpretation} \\
                            \hline
							negative	&0.62	&substantial agreement\\
							positive	&0.45	&moderate agreement\\
                            {\bf micro-average}       &{\bf 0.54}      &moderate agreement      \\

\hline
\\
\end{tabular}
\label{tab:so scotts pie}
\end{table}

\section{Conclusions}
Emotion detection and generation have a number of practical
applications \ch{including} managing customer relations, human computer interaction,
information retrieval, more natural text-to-speech systems, and in social and literary analysis.
% Even though there is work in speech and facial
% expressions, it is only recently that we see growing interest in
% analyzing text for emotions. 
However, only a small number of
limited-coverage emotion resources exist, and that too only for
English.  In this paper we show how the combined strength and wisdom
of the crowds can be used to generate a large term--emotion
association lexicon quickly and inexpensively.  This lexicon, EmoLex,
has entries for more than 10,000 word--sense pairs.  Each entry lists
the association of the a word--sense pair with 8 basic emotions.  We
used Amazon's Mechanical Turk as the crowdsourcing platform.  
% Lexicons can be created, in a similar manner, for other languages too, as long
% as there are enough Turkers who speak the target language.

We outlined various challenges associated with crowdsourcing the
creation of an emotion lexicon (many of which apply to other language
annotation tasks too), and presented various solutions to address
those challenges.  Notably, we used automatically generated word
choice questions to detect and reject erroneous annotations and to
reject all annotations by unqualified Turkers and those who indulge in
malicious data entry.  The word choice question is also an effective
and intuitive way of conveying the sense for which emotion annotations
are being requested. 

% \ch{It should} be noted that unlike certain other annotation tasks,
% there is no correct answer to many of the term--emotion questions
% given to the annotators. However, the results from the annotations
% show that for a large number of word--emotion pairs, at least four out of five
% annotators agreed on the nature of the association.
% We provide agreement values for each of the term--emotion pairs,
% allowing downstream applications to selectively use the lexicon.
We compared a subset of our lexicon with existing gold standard data
to show that the annotations obtained are indeed of high quality.  
% A detailed analysis of the lexicon revealed insights into how prevalent
% emotion bearing terms are among common unigrams and bigrams. 
We identified which emotions tend to be evoked simultaneously by the same
term, and also how frequent the emotion associations are in high-frequency words.  
We also compiled a list of 826 terms
that are not merely associated with emotions, but also refer directly to
emotions. All of the 10,170 terms in the lexicon are also annotated
with whether they have a positive, negative, or neutral semantic
orientation.

\section{Future Directions}
Our future work includes expanding the coverage of the lexicon even
further, creating similar lexicons in other languages, % such as German,
identifying cross-cultural and cross-language differences in emotion
associations, and using the lexicon in
various emotion detection applications such as those listed in Section~\ref{sec: apps}.
\cite{MohammadY11} describe some of these efforts, in which we use the {\it Roget's
Thesaurus} as the source of target terms, and create an emotion lexicon with
entries for more than 24,000 word--sense pairs (covering about 14,000 unique word-types).
% The lexicons will be made available for download.\footnote{http://www.purl.org/net/emolex} 
 We will use this manually created emotion lexicon
to evaluate automatically generated lexicons, such as the polarity
lexicons by \cite{Turney03} and \cite{MohammadDD09}.  We will explore
the variance in emotion evoked by near-synonyms, and also how common
it is for words with many meanings to evoke different emotions in
different senses.

We are interested in further improving the annotation process by
applying {\ebf Maximum Difference Scaling} (or {\ebf MaxDiff})
\cite[]{Louviere91,LouviereF92}.  In MaxDiff, instead of asking
annotators for a score representing how strongly an item is associated
with a certain category, the annotator is presented with four or five
items at a time and asked which item is {\it most} associated with the
category and which one the {\it least}. The approach forces annotators to
compare items directly, which leads to better annotations
\cite[]{LouviereF92,CohenC03}, which we hope will translate into
higher inter-annotator agreements.  Further, if $A, B, C,$ and $D$ are the
four items in a set, by asking only the most and least questions, we
will know five out of the six inequalities. For example, if $A$ is the
maximum, and $D$ is the least, then we know that $A > B, A > C, A > D, B
> D, C > D$. This makes the annotations significantly more efficient
than just providing pairs of items and asking which is more associated
with a category. Hierarchical Bayes estimation can then be used to
convert these MaxDiff judgments into scores (from 0 to 10 say) and to
rank all the items in order of association with the category.

% In many ways, 
% work on emotion analysis is related to work on detecting
% polarity, and 
Many of the challenges associated with polarity analysis
have correspondence in emotion analysis too. 
% \ch{Certain} key challenges still remain for both areas:
For example, using context information in addition to prior probability
of a word's polarity or emotion association, to determine the true
emotional impact of a word in a particular occurrence.
Our emotion annotations are at word-sense level, yet accurate word sense
disambiguation systems must be employed to make full use of this information.
For example, \cite{Rentoumi09} show that word sense disambiguation improves detection
of polarity of sentences.
There is also a need for algorithms to identify who is experiencing an emotion, and
determine what or who is evoking that emotion.  Further, given a sentence or a
paragraph, the writer, the reader, and the entities
mentioned in the text may all have different emotions associated with
them.  Yet another challenge is how to handle negation of emotions.
For example, {\it not sad} does not usually mean {\it happy}, whereas {\it not
happy} can often mean {\it sad}.

\ch{Finally}, emotion detection can be used as a tool for social and literary analysis.
For example, how have books portrayed different entities over time?
Does the co-occurrence of fear words with entities (for example, cigarette, or homosexual,
or nuclear energy) reflect the feelings of society as a whole towards these entities?
What is the distribution of different emotion words in novels and plays?
How has this distribution changed over time, and across different genres?
Effective emotion analysis can help identify
trends and lead to a better understanding of humanity's changing perception of the world around it.

% We are also interested din developing effective visualizations for emotions in text.
% Apart from basic
% We will make the lexicons publicly available for free download.

\section*{Acknowledgments}
This research was funded by the National Research Council Canada
(NRC).  We are grateful to the reviewers for their thoughtful
comments.  Thanks to Joel Martin, Diana Inkpen, and Diman Ghazi for 
discussions and encouragement. Thanks to Norm Vinson and the Ethics Committee at NRC for
examining, guiding, and approving the survey.  And last but not least,
thanks to the more than 2000 anonymous annotators who answered the
emotion survey with diligence and care.

\label{lastpage}

\bibliographystyle{natbib}
\bibliography{references}

\end{document}